\documentclass[mathfont=cm, accepted]{uai2021}
\usepackage[round]{natbib}
\usepackage{amsmath,bm} 
\usepackage{amssymb}  
\usepackage{url}
\usepackage{times}
\usepackage{float}
\usepackage{graphics} 
\usepackage{epsfig} 
\usepackage{xcolor}
\usepackage{hyperref}
\usepackage{caption}
\usepackage{subcaption}
\usepackage{algorithm}
\usepackage[noend]{algpseudocode}
\usepackage{booktabs}
\usepackage{multirow}

\title{On the Effects of Quantisation on Model Uncertainty in Bayesian Neural Networks}

\author[1]{\href{mailto:Martin Ferianc <martin.ferianc.19@ucl.ac.uk>?Subject=Quantised BNN Paper}{Martin Ferianc}{}}
\author[2]{Partha Maji}
\author[3]{Matthew Mattina}
\author[1]{Miguel Rodrigues}
\affil[1]{%
    Electronic and Electrical Engineering Department\\
    University College London\\
    London, UK
}
\affil[2]{%
    Arm ML Research Lab\\
    Cambridge, UK
}
\affil[3]{%
    Arm ML Research Lab\\
    Boston, USA
}

\begin{document}

\maketitle

\begin{abstract}
Bayesian neural networks (BNNs) are making significant progress in many research areas where decision-making needs to be accompanied by uncertainty estimation. Being able to quantify uncertainty while making decisions is essential for understanding when the model is over-/under-confident, and hence BNNs are attracting interest in safety-critical applications, such as autonomous driving, healthcare, and robotics.
Nevertheless, BNNs have not been as widely used in industrial practice, mainly because of their increased memory and compute costs. In this work, we investigate quantisation of BNNs by compressing 32-bit floating-point weights and activations to their integer counterparts, that has already been successful in reducing the compute demand in standard pointwise neural networks. We study three types of quantised BNNs, we evaluate them under a wide range of different settings, and we empirically demonstrate that a uniform quantisation scheme applied to BNNs does not substantially decrease their quality of uncertainty estimation.
\end{abstract}
\section{Introduction}\label{sec:introduction}
Bayesian neural networks (BNNs) can describe complex stochastic
patterns by treating their weights as learnable random variables that provide well-calibrated uncertainty estimates~\citep{neal1993bayesian, ghahramani2015probabilistic, blundell2015weight, gal2016dropout, chen2014stochastic}. In addition to modelling uncertainty, by treating a neural network (NN) through Bayesian inference, it gains robustness to over-fitting thereby offering the means to leverage small data pools~\citep{ghahramani2015probabilistic}.

BNNs have become relevant in practical applications where the quantification of uncertainty is essential such as in medicine~\citep{liang2018bayesian}, autonomous driving~\citep{mcallister2017concrete} or risk assessment~\citep{mackay1995bayesian}. Nevertheless, Bayesian models come with a prohibitive
computational cost during evaluation~\citep{gal2016dropout, blundell2015weight}. While evaluating, it is analytically intractable to compute the posterior prediction. Hence, most methods approximate the posterior through Monte Carlo (MC) sampling~\citep{gal2016dropout,blundell2015weight,chen2014stochastic}, which depends on multiple feed-forward runs through the BNNs and optionally random number generation. 

In contrast to pointwise NNs, that are increasingly used for applications on the edge, the computational cost associated with BNNs currently prevents their use on resource-constrained platforms. These platforms exhibit smaller memory and lower compute capabilities involving 8-bit integer arithmetic. Quantisation has been widely used in pointwise NNs~\citep{jacob2018quantization, choukroun2019low, krishnamoorthi2018quantizing} to lower their compute demand and make them more compatible with edge devices. In quantisation, floating-point representation is reduced to an integer representation, which enables substantial resource savings in practical applications. By quantising weights and activations of pointwise NNs to 8-bit integers, it is possible to achieve up to 4$\times$ improvements in latency with a quarter of the original memory footprint of the baseline 32-bit floating-point implementation~\citep{guo2017survey,jacob2018quantization}. Nevertheless, there has not been a comprehensive study into whether BNNs could attain the same hardware benefits under quantisation and whether it impacts their predictive accuracy or uncertainty.

In this work, we study quantisation of BNNs based on three widely adapted Bayesian inference schemes: Monte Carlo Dropout~\citep{gal2016dropout}, Bayes-By-Backprop~\citep{blundell2015weight} and Stochastic Gradient Langevin Dynamics with Hamiltonian Monte Carlo~\citep{chen2014stochastic}. Furthermore, we investigate the effect of quantisation of both weights and activations of BNNs using different integer representations through quantisation aware training. Our main contributions
are two-fold \textit{1)} Methodology for uniform quantisation of three different types of Bayesian inference; \textit{2)} An empirical demonstration that lowering arithmetic precision of weights and activations from 32-bit floating-point to $\leq$8-bit integers does not substantially detriment accuracy and uncertainty estimation quality of Bayesian neural networks across different datasets, network architectures and tasks. To the best of our knowledge, we are the first ones to attempt an empirical investigation in this direction with respect to widely compared and accessible benchmarks. 
The code is available at \url{https://git.io/JtSJG}.


\section{Preliminaries and Related Work}\label{sec:preliminaries_related_work}
 In this Section we review Bayesian learning, quantisation of neural networks and related work.

\subsection{Bayesian Neural Networks}\label{sec:background_bnns}
The aim of Bayesian inference is to learn the distribution over the weights of the BNN $\boldsymbol{w}$ with respect to some training dataset of tuples $\mathcal{D}=\{(\boldsymbol{x}_n, \boldsymbol{y}_n)_{n=1}^N\}$, where $\boldsymbol{x}_n$ are the inputs and $\boldsymbol{y}_n$ are the associated targets. Given the belief about the noise in the data in the shape of the likelihood $p(\boldsymbol{y}| \boldsymbol{x}, \boldsymbol{w})$ and the prior distribution over weights $p(\boldsymbol{w})$, they come together under the Bayes rule $p(\boldsymbol{w}| \boldsymbol{x}, \boldsymbol{y}) = \frac{p(\boldsymbol{y}| \boldsymbol{x}, \boldsymbol{w})p(\boldsymbol{w})}{p(\boldsymbol{y}| \boldsymbol{x})}$. Nevertheless, due to the high dimensionality of BNN it is intractable to compute the posterior $p(\boldsymbol{w}| \boldsymbol{x}, \boldsymbol{y})$ and it needs to be approximated with respect to $q(\boldsymbol{w} | \boldsymbol{\theta}, \boldsymbol{x}, \boldsymbol{y})$ and some learnable parameters $\boldsymbol{\theta}$.
The resultant distribution $q(.)$ can then be used to make predictions for previously unseen data $\boldsymbol{x}^*, \boldsymbol{y}^*$ through an integral $p(\boldsymbol{y}^* | \boldsymbol{x}^*) = \int p(\boldsymbol{y}^* | \boldsymbol{x}^*, \boldsymbol{w})q(\boldsymbol{w} | \boldsymbol{\theta}, \boldsymbol{x}, \boldsymbol{y}) d\boldsymbol{w}$. This integral is again intractable due to the posterior and it needs to be approximated through through MC sampling with $L$ samples as: $
p(\boldsymbol{y}^*|\boldsymbol{x}^*) = \frac{1}{L} \sum_{l=1}^L p(\boldsymbol{y}^*|\boldsymbol{x}^*, \boldsymbol{w}_l); \boldsymbol{w}_l \sim q(\boldsymbol{w} | \boldsymbol{\theta}, \boldsymbol{x}, \boldsymbol{y})
$.
The sampling procedure requires efficient processing to reduce the compute cost of the forward pass through the BNN $L$ times. In this work, we approach this challenge through investigating quantisation applied to BNNs' weights and activations in order to enable their efficient processing.

\subsection{Quantisation} Reduction in bit-width precision~\citep{jacob2018quantization,krishnamoorthi2018quantizing,choukroun2019low} has demonstrated significant benefits in lowering the resource consumption of pointwise NNs in hardware. In quantisation, 32-bit floating-point representations of weights, and optionally, activations are reduced to an integer, usually 8-bit representation, which enables substantial savings in memory and compute resources in real-world applications. This helps to reduce energy consumption and improve inference speed. 
If the quantisation happens after training, it is called post-training quantisation. If it happens with an additional training with fewer iterations and much smaller learning rate after the main portion of the inference, it is called quantisation aware training (QAT)~\citep{jacob2018quantization}. By using QAT, practitioners have observed smaller accuracy drop in the quantised model~\citep{jacob2018quantization}, compared to post-training quantisation. The support of only integer arithmetic in hardware has two main outcomes: (1) decrease in size of the required memory and the complexity of the hardware to perform the computation; (2) decrease in latency due to the simplicity of integer computation, in comparison to floating-point~\citep{cai2018vibnn}. These benefits present a strong case for investigating quantisation of BNNs.

\subsection{Related Work} Only recently, there appeared works outside of the realm of pointwise NNs that interconnected Bayesian thinking with quantisation~\citep{su2019sampling, achterhold2018variational, cai2018vibnn, van2020bayesian}.

\citet{achterhold2018variational} developed a sophisticated method for quantisation and pruning for pointwise NNs, albeit by using Bayesian inference. They initially train a BNN with improper priors, constructed to be quantisation and pruning-friendly, and after training, convert it to a quantised pointwise NN. Although, the pointwise NNs can achieve significant reduction in memory consumption, the resultant non-quantised BNNs are actually unable to estimate uncertainty, due to improper priors~\citep{hron2017variational}. Similarly,~\citet{van2020bayesian} used Bayesian inference to obtain sparse quantised pointwise NNs. In VIBNN, ~\citet{cai2018vibnn} developed an efficient hardware accelerator for feed-forward BNNs trained through Bayes-by-backprop~\citep{blundell2015weight} algorithm. The authors demonstrated impressive compute resource savings, but they did not detail their quantisation scheme or its impact on the uncertainty estimation capabilities of the BNN. \citet{su2019sampling} proposed a method for learning quantised BNNs directly, where the range of the found activations and weights is limited to two integer values. They demonstrated that the uncertainty estimation can be preserved in the learned model. However, the work of~\citet{su2019sampling} does not allow quantisation of modern networks involving batch normalisation and skip-connections (e.g. ResNet). Additionally, in their scheme binary weights (-1, 1) need to be stored as real-valued parameters. Furthermore, their scheme in practice would require development of a custom hardware accelerator. Custom hardware accelerators are rarely used in real-world settings. Most emerging NPUs are optimised for fixed 8-bit integer arithmetic only. For existing low-resource scenarios in embedded and IoT applications CPUs are optimised for 8-bit arithmetic. 

In this paper we propose to learn quantised BNNs directly -- as in ~\citep{su2019sampling}. In contrast to their work, we consider a range of widely used Bayesian inference methods, without the need for changes in the method or  architectures. In detail, we focus on uniform quantisation, that is commonly supported in hardware (NPU, TPU, GPU)~\citep{krishnamoorthi2018quantizing}.

\section{Methodology}\label{sec:methodology}
In this Section we describe quantised BNNs, by first discussing the theory behind quantisation followed by its applicability to the respective Bayesian inference methods.

\subsection{Uniform Affine Quantisation}\label{sec:methodology_q}
The most light-weight quantisation method is an uniform affine mapping of 32-bit floating-point values $f$ to integers $q$~\citep{jacob2018quantization} as shown in \eqref{eq:quantisation_1}:
\begin{equation}
    f = S(q-Z)
    \label{eq:quantisation_1}
\end{equation}
where $S$ and $Z$ are the scale and the zero-point respectively, which are learnable parameters. The $S$ remains in floating-point representation and it effectively represents a quantisation bin-width, whereas the $Z$ is an integer of the same bit-width $n$ as $q$ and it represents the mapping of the value $0$. The values of $S$ and $Z$ are affected by the target $n$, which restricts their range. 

Assuming initially a standard pointwise linear layer with floating-point weights $\boldsymbol{f}_w\in \mathbb{R}^{M \times F}$, input $\boldsymbol{f}_i\in \mathbb{R}^{I \times M}$ and output $\boldsymbol{f}_o\in \mathbb{R}^{I \times F}$, where $M$ and $F$ correspond to the input and output feature size for a batch consisting of $I$ samples, the computation for their quantised counterparts $\boldsymbol{q}_w, \boldsymbol{q}_i, \boldsymbol{q}_o$ is obtained with respect to~\eqref{eq:quantisation_1} as follows: Linear output without quantisation is computed as $\boldsymbol{f}_{o}= \boldsymbol{f}_{i}\boldsymbol{f}_{w}$. Substituting each term with~\eqref{eq:quantisation_1}, we have $
S_{o}(\boldsymbol{q}_{o}-Z_{o}) =  S_{w}(\boldsymbol{q}_{w}-Z_{w})S_{i}(\boldsymbol{q}_{i}-Z_{i})
$
which can be rewritten as in~\eqref{eq:quantisation_3}:
\begin{multline}
\boldsymbol{q}_{o} = Z_{o}+\dfrac{S_{w}S_{i}}{S_{o}}(MZ_{w}Z_{i}-Z_{i}\sum q_{w}-Z_{w}\sum q_{i}+\\ \boldsymbol{q}_{w}\boldsymbol{q}_{i}) 
\label{eq:quantisation_3}
\end{multline}

The respective sums are performed first for each column for $\boldsymbol{q}_w$ and each row $\boldsymbol{q}_i$ and broadcast to the resultant matrix dimension, similarly to scalars $S$ and $Z$. Note that, the terms not involved with any $\boldsymbol{q}_i$ are independent of the input, which means they can be computed offline. Similarly, if the layer has a bias term, or it is followed by a batch normalisation (BN)~\citep{ioffe2015batch}, the BN affine parameters or bias  can be fused into the weights after the individual $S$ and $Z$ have been inferred~\citep{krishnamoorthi2018quantizing}. The same pattern can then be used to compute the output of more complicated operations, such as convolutions~\citep{jacob2018quantization}. Note that, the bit-width $n$ does not need to be the same for weights and activations.

The scale ($S$) and the zero-point ($Z$) parameters are learned by simulating quantisation through fine-tuning resulting in quantisation aware training (QAT). In this work we focus on QAT, which has been preferred to post-training quantisation since it is shown that it achieves higher accuracy, especially in smaller models~\citep{jacob2018quantization}. In the next Section we introduce QAT-based methods applied to NNs with respect to Bayesian inference.

\subsubsection{Quantisation Aware Training (QAT)}\label{sec:methodology_q_qat}
\begin{figure}
    \centering
    \includegraphics[width=0.85\linewidth]{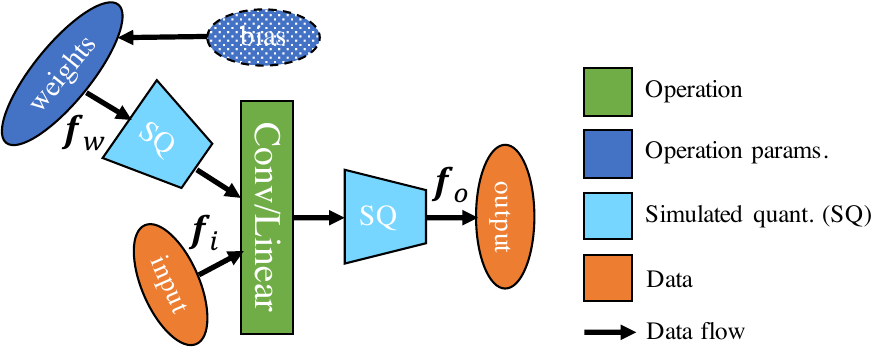}
    \caption{Fine-tuning of a standard pointwise Convolution/Linear layer with simulated quantisation (SQ). All computation is carried out using 32-bit floating-point arithmetic. SQ nodes are injected into the computation to simulate the effects of quantisation. After fine-tuning, the SQ modules are removed and the weights, with folded in bias, and computation are quantised. In a quantised regime, floating-point data $\boldsymbol{f}$ are replaced by $\boldsymbol{q}$.}
    \label{fig:qat}
\end{figure}
QAT is achieved by simulating quantisation effects in the forward pass of training, while backpropagation and all weights are represented in floating-point~\citep{jacob2018quantization}. The simulation is achieved by implementing rounding behaviour, that can be hardware platform specific, while performing floating-point arithmetic and then using a straight through gradient estimator~\citep{bengio2013estimating} in the backward pass. 
\begin{itemize}
    \item Weights' quantisation is simulated prior to being combined with the input, to avoid dynamic quantisation during runtime.
    \item Activation or operation output quantisation is simulated at points where they would be during inference - after the activation function is applied or after addition or concatenation of outputs of several layers as in ResNets~\citep{jacob2018quantization, he2016deep}.
\end{itemize}
Concretely in this work, we adopt the element-wise quantisation function as shown by~\citet{jacob2018quantization} for all tensors individually, and we assume hardware fusion of the common ReLU activation, BN and bias into the operation as done in practice~\citep{krishnamoorthi2018quantizing}. The quantisation and its simulation are parametrised by $n$, which is user specified, and a clamping range consisting of a minimum $a;a=\min\boldsymbol{f}$ and a maximum $b;b=\max\boldsymbol{f}$ for the given tensor. Individual $a$ and $b$ are being observed on the training and validation datasets, for each activation output and weight. To observe the most efficient clamping range bounds $a,b$, it is necessary to record the minimum and maximum values of the respective tensors during training and then individually aggregate them via exponential moving average, because of perturbations in outputs and weights due to QAT fine-tuning. The $a,b,n$ continually map to $S;S=\frac{b-a}{n-1}$ and $Z;Z=\textrm{round}(\min\frac{\boldsymbol{f}}{S})$ that are being used for the simulation and the end values are then used for the actual quantisation, following equation~\eqref{eq:quantisation_1}. The computational graph with respect to QAT is visually represented in Figure~\ref{fig:qat} and in pseudo code in Algorithm~\ref{alg:qat}. In the next Section we describe how this scheme can be used to obtain quantised BNNs. 
\begin{algorithm}
\caption{Quantisation Aware Training}
\label{alg:qat}
\begin{algorithmic}[1]
 \State Inference of a floating-point model until convergence.
 \State Fusion of biases and batch normalisation statistics with weights
 \State Insertion of simulated quantisation (SQ) modules after weights and operations' outputs.
 \State Fine-tuning, simulating quantisation and recording individual $a,b$ per tensor in the computational graph.
 \State Computation of individual $S$ and $Z$, quantisation of weights and computation of offline constants to prepare the model for integer arithmetic evaluation.
\end{algorithmic}
\end{algorithm}
\subsection{Quantised Bayesian Neural Networks}
In this work we develop schemes for performing quantisation aware training (QAT) for Bayesian inference methods for both their weights and activation outputs. Note that, we propose to use QAT exclusively after Bayesian inference, and with minimal fine-tuning, such that the parameters learned through the Bayesian inference are not compromised. 
We illustrate the quantisation process with the help of a linear layer and notation from Section~\ref{sec:methodology_q}. In general, it is necessary to only discuss the placement of SQ nodes with respect to the compute graphs and step 2. from Algorithm~\ref{alg:qat} for the respective Bayesian inference methods, following the rules introduced in the bullet-points in the previous Section. Other steps are exactly the same as for the pointwise counterpart. 
\begin{figure}
    \centering
    \includegraphics[width=1.0\linewidth]{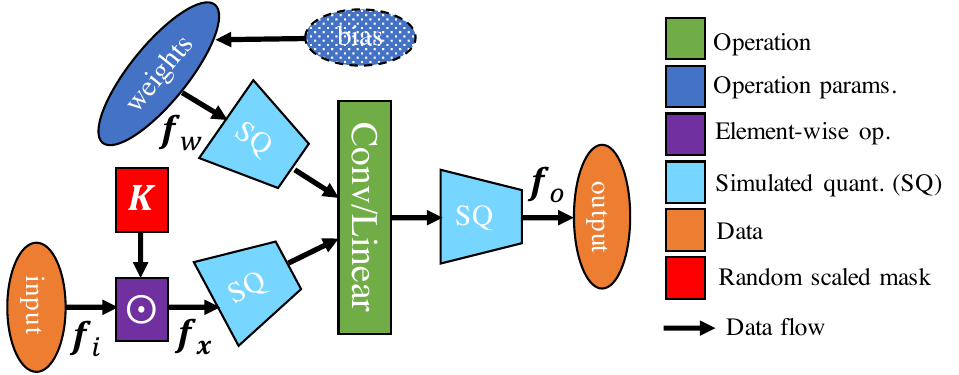}
    \caption{Quantisation for Monte Carlo Dropout.}
    \label{fig:quant_mcd}
\end{figure}
\begin{figure}
    \centering
    \includegraphics[width=1.0\linewidth]{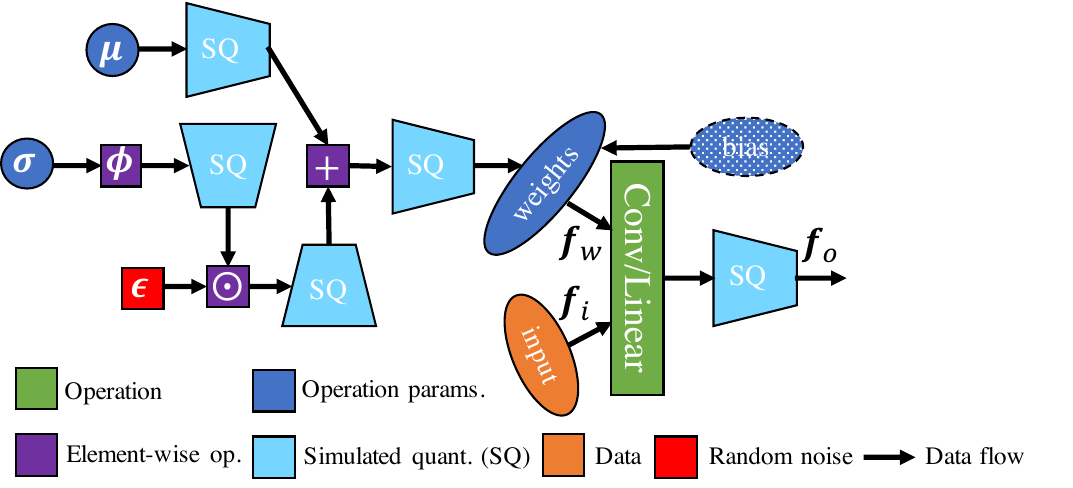}
    \caption{Quantisation for Bayes-by-Backprop.}
    \label{fig:quant_bbb}
\end{figure}
\begin{figure*}[t]
\centering
\begin{subfigure}{.5\textwidth}
  \centering
  \includegraphics[width=1.\linewidth]{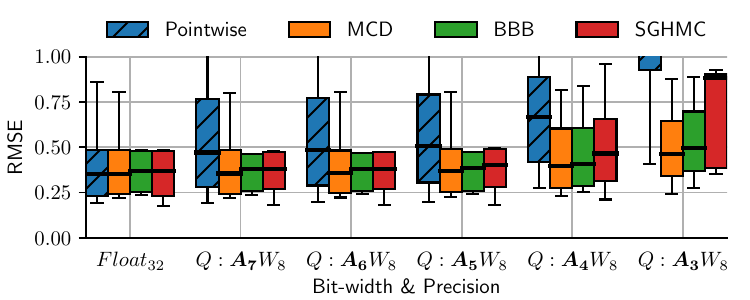}
  \caption{}
  \label{fig:regression_activation_rmse}
\end{subfigure}%
\begin{subfigure}{.5\textwidth}
  \centering
  \includegraphics[width=0.95\linewidth]{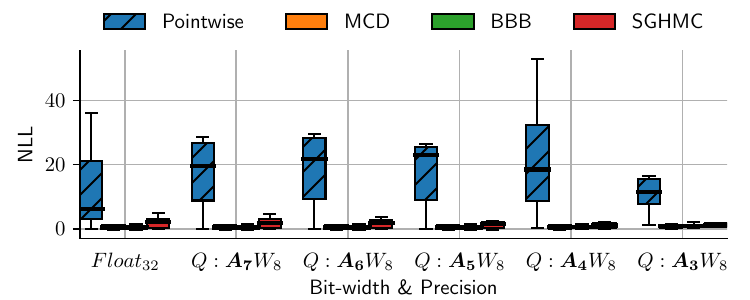}
  \caption{}
  \label{fig:regression_activation_nll}
\end{subfigure}
\caption{\textit{Changing activation precision, fixing weight precision.} Regression results with respect to root-mean-squared error (RMSE) (a) and negative log-likelihood (NLL) (b) on UCI datasets. Q stands for quantised activations (A) and weights (W). Subscript denotes bit-width. Pointwise and SGHMC collapse when the bit-width $\leq 3$ for A.}
\label{fig:regression_activation}
\end{figure*}
\begin{figure*}[t]
\centering
\begin{subfigure}{.5\textwidth}
  \centering
  \includegraphics[width=1.\linewidth]{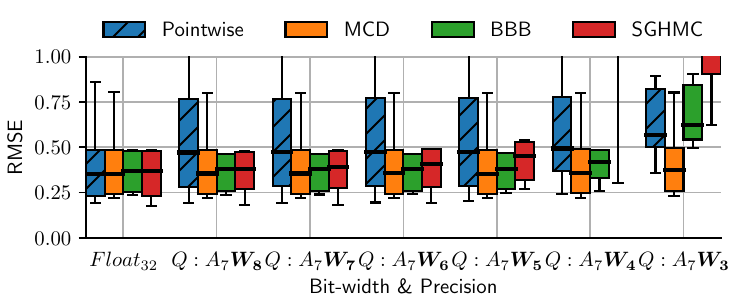}
  \caption{}
  \label{fig:regression_weight_rmse}
\end{subfigure}%
\begin{subfigure}{.5\textwidth}
  \centering
  \includegraphics[width=0.95\linewidth]{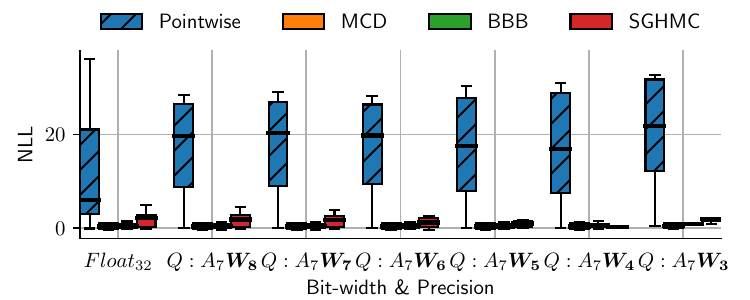}
  \caption{}
  \label{fig:regression_weight_nll}
\end{subfigure}
\caption{\textit{Fixing activation precision, changing weight precision.} Regression results with respect to root-mean-squared error (RMSE) (a) and negative log-likelihood (NLL) (b) on UCI datasets. Q stands for quantised activations (A) and weights (W). Subscript denotes bit-width. SGHMC collapses when the bit-width $\leq 4$ for W.}
\label{fig:regression_weight}
\end{figure*}
\subsubsection{Monte Carlo Dropout (MCD)} The quantisation of MCD in shown in Figure~\ref{fig:quant_mcd}. We propose methodology for quantisation of the standard MCD implementation~\citep{gal2016dropout}, that corresponds to applying Bernoulli mask of zeros and ones $\boldsymbol{K} \in \mathbb{R}^{I \times M}$ with respect to an input with $M$ features and $I$ samples for each weight-bearing layer, except the input. Additionally the masked input is scaled by the proportion of zeros in the mask, such that $\boldsymbol{f}_x=\boldsymbol{K}\odot \boldsymbol{f}_i\frac{1}{1-p}$ and $\boldsymbol{q}_x=\boldsymbol{K}\odot \boldsymbol{q}_i\frac{1}{1-p}$, where $p$ is the probability of sampling zero and $\odot$ is an element-wise multiplication. Thus $\boldsymbol{f}_x,\boldsymbol{q}_x$ values replace $\boldsymbol{f}_i,\boldsymbol{q}_i$ values with respect to equation~\eqref{eq:quantisation_3}. We add separate SQ node to the multiplication of the $\boldsymbol{K}$ with the input, since due to the factor $\frac{1}{1-p}$ and zeroing-out some inputs, the respective $S$ and $Z$ will change. When generating the $\boldsymbol{K}$, we absorb the $\frac{1}{1-p}$ into the mask for efficient computation. Thus, $\boldsymbol{K}$ does not include optimisable parameters and SQ is not directly needed after $\boldsymbol{K}$. Note that during performing QAT it is necessary to generate the mask in floating-point, while in a quantised mode the $\boldsymbol{K}$ needs to take into account the $Z$ of $\boldsymbol{q}_i$. Weights are simply quantised according to equation \eqref{eq:quantisation_1} and by adding an SQ node as discussed in the Section~\ref{sec:methodology_q_qat}.

\subsubsection{Bayes-By-Backprop (BBB)} We propose QAT methodology for BBB as shown in Figure~\ref{fig:quant_bbb}. In BBB~\citep{blundell2015weight}, the distribution over the weights is modelled explicitly such that $\boldsymbol{f}_w,\boldsymbol{q}_w{\sim}\mathcal{N}(\boldsymbol{\mu}, \boldsymbol{\sigma}^2)$, with mean $\boldsymbol{\mu}\in\mathbb{R}^{M\times F}$ and variance $\boldsymbol{\sigma}^2\in\mathbb{R}^{M\times F}$ for each weight with respect to $M$ input and $F$ output features and $\mathcal{N}$ represents a Gaussian. Nevertheless, to enable backpropagation and efficient computation of the weights, the weights are sampled with respect to a Gaussian $\boldsymbol{\epsilon}{\sim}\mathcal{N}(0,1)$, such that $\boldsymbol{f}_w=\boldsymbol{\mu} + \phi(\boldsymbol{\sigma}) \odot \boldsymbol{\epsilon}$~\citep{kingma2013auto}. $\phi(.)$ constrains the output to be positive e.g.: softplus. It is necessary to add SQ nodes and observe the statistics after each operation: application of positive element-wise $\phi(.)$, addition and multiplication to obtain $\boldsymbol{f}_w$ and subsequently $\boldsymbol{q}_w$. We simulate quantisation to compute the quantisation statistics for the means $\boldsymbol{\mu}$ as well as the positive standard deviation $\phi(\boldsymbol{\sigma})$. We do this to avoid dynamic quantisation during run-time. The quantisation for the standard deviation is performed after $\phi(.)$, which eventually bypasses the non-linearity, when quantised, and reduces the numerical errors induced by the reduced representation. Practically, this means that we do not have to perform $\phi(.)$ and there is no floating-point computation at runtime. Note that, depending on the regime, it is necessary generate $\boldsymbol{\epsilon}$ in floating-point or quantised. We found quantised $\boldsymbol{\epsilon}$ with respect to $S_\epsilon=0.0236$ and $Z_\epsilon=0$ to be performing well across different $n$ and experiments. Due to the proposed scheme, there are no changes necessary to be made with respect to equation~\eqref{eq:quantisation_3}. We avoid the computation of the gradients with respect to the ELBO's regulariser~\citep{blundell2015weight} during QAT. We found it practically difficult to perform the quantisation with respect to the non-linear computation of KL divergence (taking log, square, division).

\subsubsection{Stochastic Gradient Langevin Dynamics with Hamiltonian Monte Carlo (SGHMC)} In comparison to the previous two methods, in SGHMC~\citep{chen2014stochastic}, there is no sampling of random variables during evaluation. \citet{chen2014stochastic}, following~\citep{welling2011bayesian}, demonstrated that by adding the right amount of Gaussian noise to a standard stochastic gradient optimisation algorithm, it is possible to collect weights $\boldsymbol{w}_l$ from $l=1,\ldots,L$ several distinct optimisation steps, that can then be used to approximate the samples from the true posterior distribution over the BNN weights. Therefore, we propose to quantise each of the weight samples separately through QAT, similarly to a pointwise approach, as shown in Figure~\ref{fig:qat}. The SQ nodes are applied to each set of weight samples $l$ as well as the corresponding outputs. Thus, we propose to fine-tune each pre-trained network sample $l$ separately. 

 In all instances, QAT was used with a very small learning rate for 10 epochs. However, it is crucial that Bayesian inference was adhered to in the main phase with a bigger learning rate and substantially more epochs. 

\section{Experiments}\label{sec:experiments}
In this Section we present the tasks, datasets and their corresponding NN architectures, metrics and the implementation, followed by the observations. 

\begin{figure*}[t]
\centering
\begin{subfigure}{.5\textwidth}
  \centering
  \includegraphics[width=1.\linewidth]{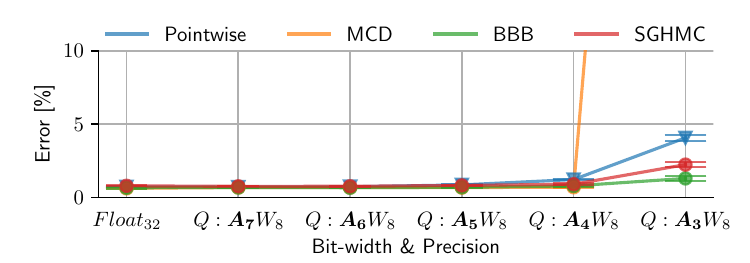}
  \caption{}
  \label{fig:mnist_activation_error}
\end{subfigure}%
\begin{subfigure}{.5\textwidth}
  \centering
  \includegraphics[width=1.\linewidth]{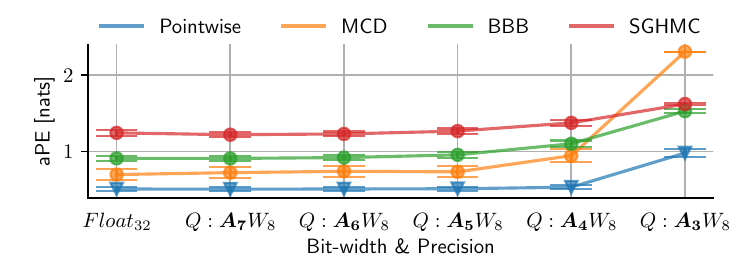}
  \caption{}
  \label{fig:mnist_activation_entropy}
\end{subfigure}
\caption{\textit{Changing activation precision, fixing weight precision.} MNIST results with respect to classification error on test data (a) and average predictive entropy (aPE) on FashionMNIST (b). Q stands for quantised activations (A) and weights (W). Subscript denotes bit-width. MCD collapses when the bit-width $\leq 3$ for A.}
\label{fig:mnist_activation}
\end{figure*}
\begin{figure*}[t]
\centering
\begin{subfigure}{.5\textwidth}
  \centering
  \includegraphics[width=1.\linewidth]{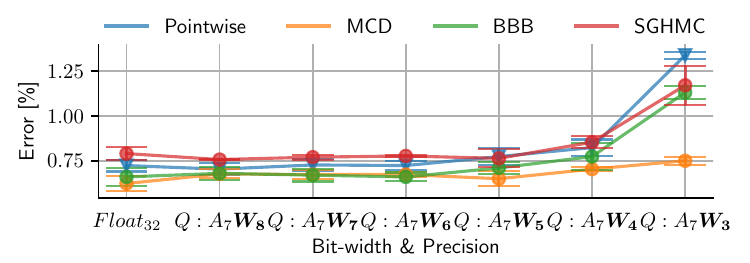}
  \caption{}
  \label{fig:mnist_weight_error}
\end{subfigure}%
\begin{subfigure}{.5\textwidth}
  \centering
  \includegraphics[width=1.\linewidth]{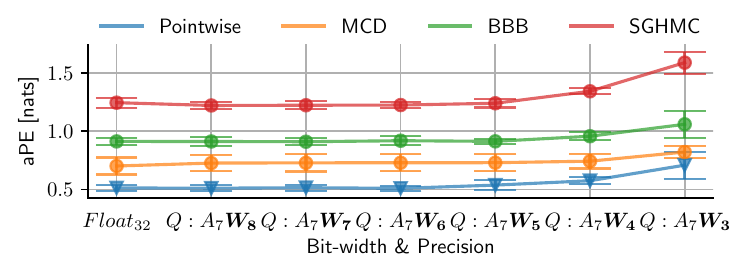}
  \caption{}
  \label{fig:mnist_weight_entropy}
\end{subfigure}
\caption{\textit{Fixing activation precision, changing weight precision.} MNIST results with respect to classification error on test data (a) and average predictive entropy (aPE) on FashionMNIST (b). Q stands for quantised activations (A) and weights (W). Subscript denotes bit-width.}
\label{fig:mnist_weight}
\end{figure*}

We consider two classes of problems: \textit{1)} regression and \textit{2)} classification. We evaluate the networks on sample datasets of tuples $\mathcal{D}$. For regression the target $y_n$ is assumed to be a real-valued $y_n\in \mathbb{R}^1$, while for classification the target $\boldsymbol{y}_n$ is a one-hot encoding of $k=1,\ldots,K$ classes such that $\boldsymbol{y}_n\in \mathbb{R}^K$. Given the input features $\boldsymbol{x}_n$, we use a BNN to model the probabilistic predictive distribution $p_w(y_n|\boldsymbol{x}_n)$ over the targets with respect to some model defined by weights $\boldsymbol{w}$, where the mean and the variance are approximated with respect to $L$ samples as $\mu_w(\boldsymbol{x}_n)=\mathbb{E}[\frac{1}{L}\sum_{l=1}^L p_{w_l}(y_n|\boldsymbol{x}_n)]$ and $\sigma^2_w(\boldsymbol{x}_n)=\mathbb{E}[\frac{1}{L}\sum_{l=1}^L(p_{w_l}(y_n|\boldsymbol{x}_n)-\mu_w(\boldsymbol{x}_n))^2]$.

For the regression we consider 
UCI datasets (housing, concrete, energy, power, wine, yacht) whereas for classification we consider classifying MNIST digits and CIFAR-10 image datasets. We used a mixture of real data to control the complexity of the experiments and observe whether it affects the uncertainty estimation quality in a quantised regime. For the regression problem we consider an architecture with an input layer followed by 3 hidden layers with 100 nodes, each followed by a ReLU activation. For MNIST we implement the common LeNet-5~\citep{lecun1998gradient}, while for CIFAR-10 we implement ResNet-18~\citep{he2016deep} with BN and skip-connections enabled. Similarly to the datasets, we chose NN architectures of increasing complexity to explore how the uncertainty estimation is impacted by trailing quantisation errors coming from a reduced precision and deeper architectures. We considered image augmentations: rotation, brightness and horizontal shift and confusion datasets: FashionMNIST for MNIST and SVHN for CIFAR-10 experiments to measure the level of uncertainty on distant or shifted datasets. The hyperparameters for all experiments were hand-tuned with reference to validation error.

From the quantisation point of view, we focus on quantisation of both weights and activations to improve on-device storage as well as computational efficiency. We considered $3\leq n \leq 8$ for weights (W) and $3\leq n \leq 7$ for activations (A) for all the proposed methods (MCD, BBB, SGHMC) and a standard pointwise implementation. We considered 1 bit lower precision for activations than for weights to avoid instruction overflow on our system. All experiments were repeated 3 times and we set $L=20$ for all methods. 
The code is available at \url{https://git.io/JtSJG}. Additional experiments and observations are in the appendix.
\begin{figure*}
\centering
\begin{subfigure}{.5\textwidth}
  \centering
  \includegraphics[width=1.\linewidth]{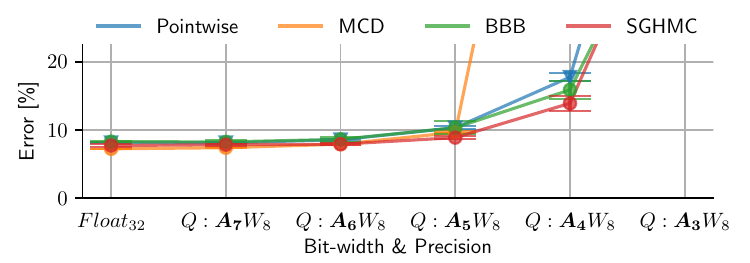}
  \caption{}
  \label{fig:cifar_activation_error}
\end{subfigure}%
\begin{subfigure}{.5\textwidth}
  \centering
  \includegraphics[width=1.\linewidth]{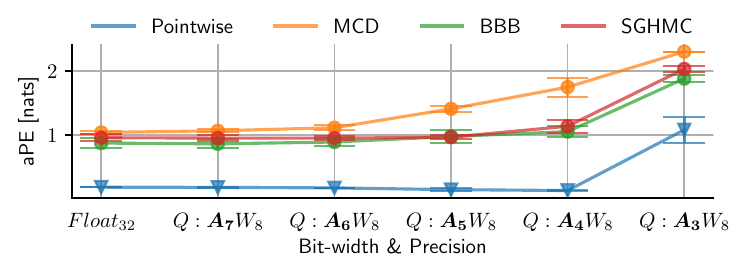}
  \caption{}
  \label{fig:cifar_activation_entropy}
\end{subfigure}
\caption{\textit{Changing activation precision, fixing weight precision.} CIFAR-10 results with respect to classification error on test data (a) and average predictive entropy (aPE) on SVHN (b). Q stands for quantised activations (A) and weights (W). Subscript denotes bit-width. All methods collapse when the bit-width $\leq 4$ for A.}
\label{fig:cifar_activation}
\end{figure*}
\begin{figure*}
\centering
\begin{subfigure}{.5\textwidth}
  \centering
  \includegraphics[width=1.\linewidth]{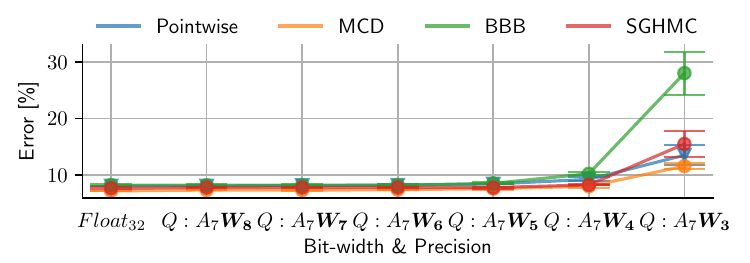}
  \caption{}
  \label{fig:cifar_weight_error}
\end{subfigure}%
\begin{subfigure}{.5\textwidth}
  \centering
  \includegraphics[width=1.\linewidth]{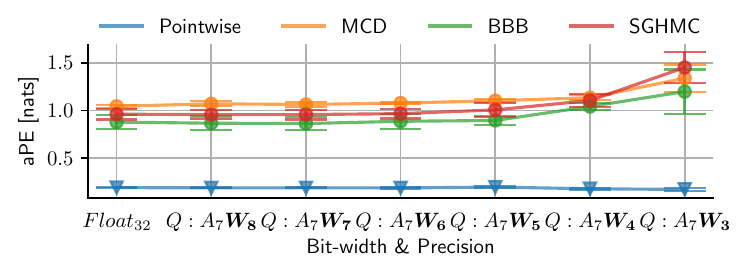}
  \caption{}
  \label{fig:cifar_weight_entropy}
\end{subfigure}
\caption{\textit{Fixing activation precision, changing weight precision.} CIFAR-10 results with respect to classification error on test data (a) and average predictive entropy (aPE) on SVHN (b). Q stands for quantised activations (A) and weights (W). Subscript denotes bit-width. All methods collapse when the bit-width $\leq 3$ for W.}
\label{fig:cifar_weight}
\end{figure*}
\vspace{-2em}
\subsection{Regression}
The results for regression for the respective methods under quantisation are presented in Figures~\ref{fig:regression_activation} (a,b) and~\ref{fig:regression_weight} (a,b). We were measuring the root-mean-squared error (RMSE) and negative log-likelihood (NLL). Every box-plot is with respect to the UCI datasets and means of 10-fold cross validation that has been done with respect to independent models. First, examining the results for changing activation precision in Figure~\ref{fig:regression_activation} (a), it can be seen from the RMSE that the Bayesian methods are more robust towards quantisation and they are able to maintain their accuracy, while a pointwise NN tends to lose its generalisability the quickest, even though it was initially marginally the most accurate. At the same time, the Bayesian inference methods are able to maintain their uncertainty estimation capabilities which can be seen in the NLL plots in Figures~\ref{fig:regression_activation}, ~\ref{fig:regression_weight} (b). Second, results plotted in Figure~\ref{fig:regression_weight} for changing weight precision and keeping the activation precision fixed, further solidify the previous observations. Nevertheless, the rate of change of the error with respect to quantisation of weights is slower in comparison to changing the activation precision. However, in both plots we notice that SGHMC is more affected by quantisation, especially weight quantisation. The weights' distributions for SGHMC for the different layers are more spread than the other 2 methods and uniform quantisation with respect to such a low precision for either weights or activations is unable to capture them. 
\vspace{-1em}
\subsection{Classification}
In this Section we present the main results with respect to evaluation on MNIST and CIFAR-10 datasets. We focused on measuring classification error, expected calibration error (ECE)~\citep{guo2017calibration} with respect to 10 bins and average predictive entropy (aPE). \textit{Further results with respect to other metrics can be seen in the appendix.}

\subsubsection{MNIST}
The results for MNIST evaluation with respect to quantised BNNs are presented in Figures~\ref{fig:mnist_activation} (a,b) and Figures~\ref{fig:mnist_weight} (a,b). In general, the results follow the same trends as demonstrated in the regression results. Nevertheless, as seen in the classification error for changing activation precision in Figure~\ref{fig:mnist_activation} (a) the respective methods are more sensitive towards changing activation precision than weight precision in comparison to results in Figure~\ref{fig:mnist_weight} (a), in particular for MCD. The scaling factor that is applied during the MCD $(\frac{1}{1-p})$ distorts the activation distribution and results in a collapse of MCD if the bit-width for the activation is too small. However, the error of the BNNs increases marginally slower than for the pointwise NN. Nevertheless, as the error increases, the predictive entropy increases as well, which can be seen in both Figures~\ref{fig:mnist_activation} (b) and \ref{fig:mnist_weight} (b) as a result their ECE also decreases. This means that quantisation has actually a regularising effect as with reduced precision for weights or activations the representation capabilities of NNs is limited and their confidence decreases. Interestingly for the collapsed MCD, this results in a complete and rightful uncertainty on the confusion dataset or the test set as seen in Figure~\ref{fig:mnist_activation} (b). These results translate also to measuring aPE and ECE on the test data, except for the pointwise control.

\begin{figure*}[t]
\begin{subfigure}{.49\textwidth}
    \centering
    \includegraphics[width=1\linewidth]{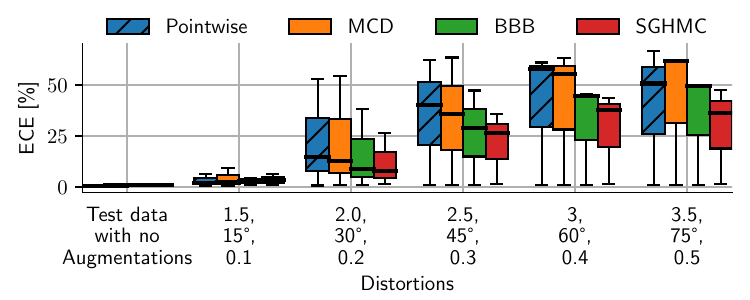}
    \caption{}
\end{subfigure}
\begin{subfigure}{.49\textwidth}
    \centering
    \includegraphics[width=1\linewidth]{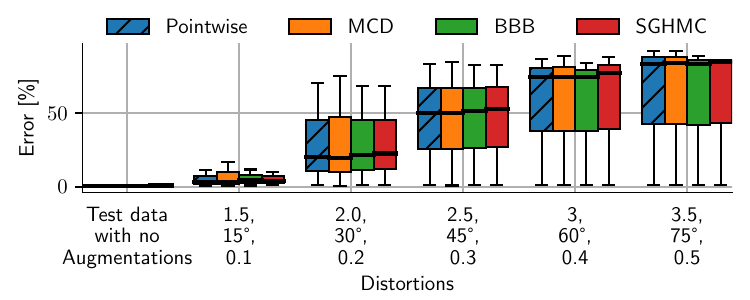}
    \caption{}
\end{subfigure}
\newline
\begin{subfigure}{.49\textwidth}
    \centering
    \includegraphics[width=1\linewidth]{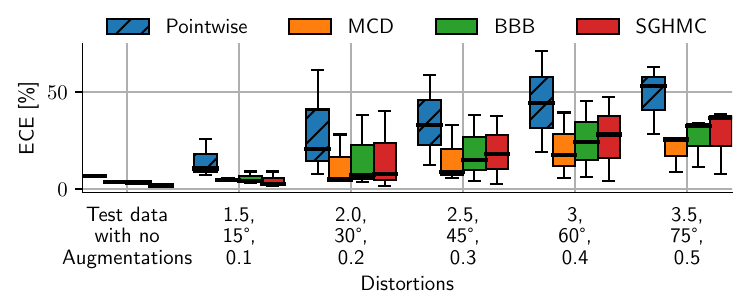}
    \caption{}
\end{subfigure}
\begin{subfigure}{.49\textwidth}
    \centering
    \includegraphics[width=1\linewidth]{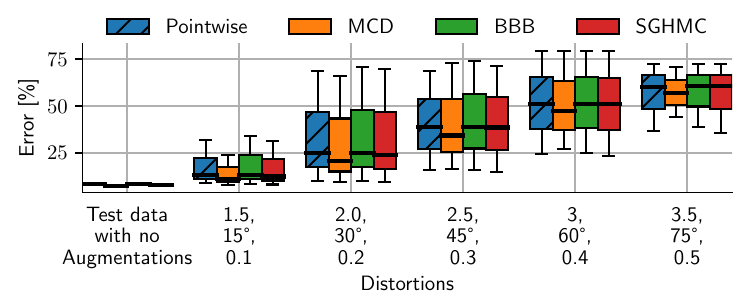}
    \caption{}
\end{subfigure}
\caption{Expected calibration error (ECE) and classification error with respect to 7-bit activations and 8-bit weights and three augmentations applied to the LeNet-5 on MNIST test set (a) and (b) and ResNet-18 on CIFAR-10 test set (c) and (d). Augmentations were: Brightness [1.5-3.5], Rotation [15°-75°] and Horizontal shift [0.1-0.5 of image size].}
\label{fig:augmentations}
\end{figure*}

In Figures~\ref{fig:augmentations} (a,b) we detail results with respect to augmentations and 7-bit quantisation of the activations and 8-bit quantisation of the weights. It can be seen that the Bayesian inference methods remain to be robust towards domain shift even under quantisation and they record marginally smaller ECE and error than the pointwise control. 

\subsubsection{CIFAR-10}
The results for CIFAR-10 with respect to quantised BNNs are presented in Figures~\ref{fig:cifar_activation} (a,b) and Figures~\ref{fig:cifar_weight} (a,b). In this experiment the differences between the Bayesian methods and the pointwise control are widened. Similarly to the previous experiments, the quantised BNNs are more susceptible to activation quantisation in comparison to weight quantisation, while comparing the results in Figures~\ref{fig:cifar_activation} (a) and Figures~\ref{fig:cifar_weight} (a). Moreover, the qunatised nets collapse earlier, $n\leq4$ for activations, given a more complex ResNet architecture. Nevertheless, as seen from Figures~\ref{fig:cifar_activation} (b) and Figures~\ref{fig:cifar_weight} (b) in no instance for any BNN method the uncertainty-related capability is damaged by quantisation, as the trends are clearly upwards in terms of the predictive entropy on the confusion dataset. However, as seen in Figure~\ref{fig:cifar_weight} (b) it is the pointwise model in particular which completely overfits the training dataset and quantisation has a negative effect on its predictive entropy. Similarly to previous experiments, as the error increases, the predictive entropy increases for BNNs which can be seen in both Figures~\ref{fig:cifar_activation} (b) and Figures~\ref{fig:cifar_weight} (b) as a result their ECE also decreases.

Next, if considering the domain shift as demonstrated in Figures~\ref{fig:augmentations} (c,d) it can be observed that while the error in Figure~\ref{fig:augmentations} (d) increases, the error of the Bayesian methods increases at the same rate as in a pointwise approach. However, when further examining Figure~\ref{fig:augmentations} (c), it can be seen that the ECE increases by far less in comparison to the pointwise approach, which makes BNNs, even under quantisation, to be more robust towards domain shift.

\section{Key Takeaways}\label{sec:key_takeaways}

In this work we proposed and evaluated a practical quantisation methodology for a variety of Bayesian inference methods applied to neural networks and in this Section we discuss the key takeaways of our empirical observations.
\begin{itemize}
    \item An uniform quantisation scheme is viable for quantisation of Bayesian neural networks unless pushed to the extrema ($\leq 4$-bits for activations or weights). For the most commonly utilised 8-bit weights and activation quantisation scheme used in hardware, we did not observe any significant degradation in accuracy or quality of uncertainty estimation in Bayesian nets in comparison to their floating-point representation. Therefore from the hardware perspective, we expect the BNNs to follow trends of pointwise methods - latency potentially decreased by $2\times$ to $4\times$ depending on the underlying hardware platform and memory consumption decreased by $4\times$ if considering 8-bits. Quantisation below 8-bits would require a custom accelerator to see its benefits where the latency and memory consumption could be decreased~\citep{guo2017survey}.
    \item The quality of predictive uncertainty of Bayesian networks stays unaffected or increases as a result of quantisation. The networks stay certain on the in-domain test data and become more uncertain on confusion or domain-shifted data.
    \item The prediction error increases at a slower rate in Bayesian neural networks, as their representation is reduced in the number of bits through quantisation, than in pointwise networks unless considering extrema.
    \item Activation quantisation seemed to affect all the net types more than weight quantisation on the accuracy, predictive entropy or calibration. SGHMC was more sensitive to weight quantisation, MCD was the most sensitive to activation quantisation.
    \item In MCD random binary masks ($\boldsymbol{K}$) could be quantised to 1-bit whereas in BBB all parameters ($\boldsymbol{\mu}$, $\boldsymbol{\sigma}$ and $\boldsymbol{\epsilon}$) need to be quantised with same number of bits as in weights to maintain model accuracy. 
    \item Experiments on different datasets and tasks suggest that Bayesian nets are relatively immune to quantisation.  
    However, complex architectures (e.g. ResNet) seem to be more affected by quantisation than simpler architectures (LeNet-5) regarding their performance.
\end{itemize}
In the future work we are going to investigate more complex non-mean-field approximations for the respective Bayesian inference methods and more expressive quantisation schemes with respect to lower ($\leq 4$-bits) precision.

\section*{Acknowledgements}
This work was partially completed while Martin Ferianc was an intern at Arm and completed through continued collaboration with Arm ML Research Lab. Martin Ferianc was also sponsored through a scholarship from the Institute of Communications and Connected Systems at UCL.

\bibliography{bib}
\bibliographystyle{apalike}

\appendix
\section{Detail of Bayesian Neural Network Methods}\label{appendix:detail_bnn}
We will illustrate the complications of the different approaches for Bayesian inference with the help of a toy example: a common linear layer with an input matrix $\boldsymbol{f}_i\in \mathbb{R}^{I\times M}$ with $I$ samples and $M$ features and some weights $\boldsymbol{f}_w \in \mathbb{R}^{M \times F}$ with the output $\boldsymbol{f}_o\in \mathbb{R}^{I\times F}$ with $F$ output features such that $\boldsymbol{f}_o = \boldsymbol{f}_i\boldsymbol{f}_w$.

\paragraph{Monte Carlo Dropout} 
The concept of MCD~\citep{gal2016dropout} lays in casting dropout~\citep{srivastava2014dropout} training in NNs as approximate Bayesian inference. Dropout can be described by applying a random element-wise mask $\boldsymbol{K} \in \mathbb{R}^{I\times M}; \boldsymbol{K} \sim \textrm{Bernoulli}(p)$ of zeros and ones with probability $0\leq p \leq 1$ to the input $\boldsymbol{f}_i$ and scaling the non-zero elements by $\frac{1}{1-p}$ as $\boldsymbol{f}_o= \boldsymbol{f}_w\left(\frac{1}{1-p}\boldsymbol{K}\odot \boldsymbol{f}_i \right)$. The authors of MCD show that the use of dropout in NNs before every weight-bearing layer can be interpreted as a Bayesian approximation and by applying dropout, it can approximate the integral over the models’ weights~\citep{gal2016dropout}. Therefore, to estimate the predictive  distribution $p(\boldsymbol{y}^* | \boldsymbol{x}^*)$ it is needed to collect the results of $L$ forward passes, while sampling and applying the element-wise masks. Training is usually done through a single sample. Therefore, the only implementation-wise complications of this method are the need for random generation of zeros and ones and their subsequent element-wise application. The number of parameters, and thus memory footprint, stays constant. 

\paragraph{Bayes-By-Backprop} 
In BBB~\citep{blundell2015weight} the weight uncertainty is modelled explicitly, by assuming an approximation $q(\boldsymbol{w}| \boldsymbol{x}, \boldsymbol{y}, \boldsymbol{\theta})$ for the posterior $p(\boldsymbol{w}| \boldsymbol{x}, \boldsymbol{y})$ with respect to learnable parameters $\boldsymbol{\theta}$. The learning is performed through minimising the distance bound between $q(\boldsymbol{w}| \boldsymbol{x}, \boldsymbol{y}, \boldsymbol{\theta})$ and  $p(\boldsymbol{w}| \boldsymbol{x}, \boldsymbol{y})$. The most common approximation $q$ for weights $\boldsymbol{w}$ is a mean-field approximation, such that $\boldsymbol{w}{\sim}\mathcal{N}(\boldsymbol{\mu}, \boldsymbol{\sigma}^2)$, with individual mean $\boldsymbol{\mu}\in\mathbb{R}^{M\times F}$ and variance $\boldsymbol{\sigma}^2\in\mathbb{R}^{M\times F}$ for each weight where $\boldsymbol{\theta}=\{\boldsymbol{\mu}, \boldsymbol{\sigma}^2\}$ and $\mathcal{N}$ represents a Gaussian distribution~\citep{kingma2013auto, ranganath2014black}. \citet{kingma2013auto} have introduced the \textit{reparametrisation trick}, that allows sampling of the weights with respect to the $q$, such that $\boldsymbol{f}_w  = \boldsymbol{\mu} + \boldsymbol{\epsilon}\odot\phi(\boldsymbol{\sigma})$ where $\boldsymbol{\epsilon}{\sim}\mathcal{N}(\boldsymbol{0}, \boldsymbol{I})$, $\boldsymbol{I}$ is an identity and $\phi(.)$ is a positive-forcing function e.g. softplus. The sampled $\boldsymbol{w}$ can then be used, such that $\boldsymbol{f}_o = \boldsymbol{f}_i\boldsymbol{f}_w$. Similarly to MCD, to estimate the predictive distribution $p(\boldsymbol{y}^* | \boldsymbol{x}^*)$ it is needed to collect the results of $L$ forward passes with respect to $L$ weight samples. Training is usually done through a single sample. This method requires the ability of the hardware to efficiently sample a more complex, Gaussian distribution. Moreover, this model uses double the number of parameters for the same network size, due to the means paired with variances.

\paragraph{Stochastic Gradient Langevin Dynamics with Hamiltonian Monte Carlo}
In comparison to the previous two approaches, in SGHMC~\citep{chen2014stochastic}, it is not necessary to perform sampling and random number generation during evaluation. The $\boldsymbol{w}_l$ corresponding to a single set of weights from the ensemble can then be used directly during evaluation instead of sampling them via $\boldsymbol{w}_l{\sim}p(\boldsymbol{w} | \boldsymbol{x}, \boldsymbol{y})$ as in the case of the two previous methods. To obtain $p(\boldsymbol{y}^* | \boldsymbol{x}^*)$ it is needed to collect the results of $L$ forward passes with respect to the ensemble with $L$ members corresponding to $L$ weights $\boldsymbol{w}$. Training is performed similarly to standard pointwise NNs. In comparison to the previous two methods, this method does not require sampling of a distribution during evaluation. However, it requires $L\times$ more memory resources in comparison to pointwise NNs, to store the entire ensemble. At the same time, it is necessary to consider the extra time needed to load the weights $\boldsymbol{w}$ to memory.
\section{Metrics}

In addition to measuring the root-mean-squared error (RMSE) and the classification error, we establish metrics for the evaluation of the quantified uncertainty.
\subsection{Negative-log likelihood}
Based on~\citep{lakshminarayanan2017simple}, our base metric for evaluating the quality of the predictive uncertainty is the negative-log likelihood (NLL). We chose averages in our evaluations, due to easier interpretability and consistence. In the regression case, $\textrm{NLL}$ can be formulated with respect to a single-valued Gaussian as in equation~\eqref{eq:nll_regression}. 
\begin{multline}
  \textrm{NLL} = \frac{1}{N}\sum_{n=1}^N\frac{\log \sigma^2_w(\boldsymbol{x}_n)}{2} + \frac{(y_n-\mu_w(\boldsymbol{x_n}))^2}{2\sigma^2_w(\boldsymbol{x}_n)} \\ + \log\sqrt{2\pi}.
    \label{eq:nll_regression}
  \end{multline}
In the classificaiton case, $\textrm{NLL}$ can be formulated with respect to cross-entropy as in equation~\eqref{eq:nll_classification}. 
 \begin{equation}
  \textrm{NLL} = - \frac{1}{N}\sum_{n=1}^N\sum_{k=1}^K y_{n}^{k}\log  \mu_{w}^{k}(\boldsymbol{x_n})
    \label{eq:nll_classification}
\end{equation}
\subsection{Predictive Entropy} In case of classification for which the labels are not available, which is the case for most out-of-distribution datasets, we measure the quality of the uncertainty prediction with respect to the average predictive entropy (aPE) as in equation~\eqref{eq:entropy}.
\begin{equation}
  \textrm{aPE} = - \frac{1}{N}\sum_{n=1}^N\sum_{k=1}^K\mu_{w}^{k}(\boldsymbol{x_n})\log \mu_{w}^{k}(\boldsymbol{x_n})
    \label{eq:entropy}
\end{equation} 
\subsection{Expected Calibration Error} Additionally, we measure the calibration of the BNNs and their sensitivity through expected calibration error (ECE)~\citep{guo2017calibration}. ECE relates confidence with which a network makes predictions to accuracy, thus it measures whether a network is over-confident or under-confident in its predictions, with respect to the softmax output.
To compute ECE, the authors propose to discretize the prediction probability interval into a fixed number of bins, and assign each probability to the bin that encompasses it. The ECE is the difference between the fraction of predictions in the bin that are correct (accuracy) and the mean of the probabilities in the bin (confidence). ECE computes a weighted average of this error across bins as shown in equation~\eqref{eq:ece}, where $n_b$ is the number of predictions in bin $b$ and accuracy($b$) and confidence($b$) are the accuracy and confidence of bin $b$, respectively. We set $B=10$.
\begin{equation}
  \textrm{ECE} = \sum_{b=1}^{B}\frac{n_{B}}{N}| \text{accuracy}(b) - \text{confidence}(b) |
    \label{eq:ece}
\end{equation}
To summarise, for regression we are measuring RMSE and $\textrm{NLL}$ and for image classification problems we were observing the classification error, NLL, aPE and ECE.

\section{Additional Results}

In this Section we report additional measurements with respect to test data, confusion data or the domain shifts for the image classification problems.

\subsection{MNIST}

\begin{figure*}
\begin{subfigure}{.49\textwidth}
  \centering
  \includegraphics[width=1.\linewidth]{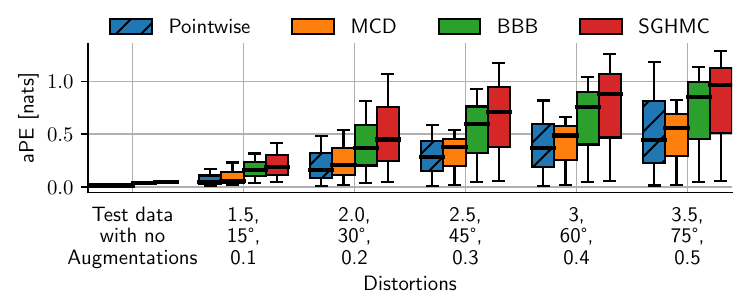}
  \caption{}
  \label{fig:appx_mnist_augmentations_entropy}
\end{subfigure}
\centering
\begin{subfigure}{.49\textwidth}
  \centering
  \includegraphics[width=1.\linewidth]{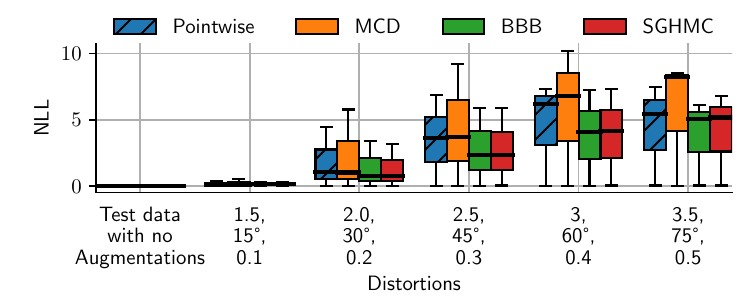}
  \caption{}
  \label{fig:appx_mnist_augmentations_nll}
\end{subfigure}
\caption{Average predictive entropy (aPE) (a) and negative log-likelihood (NLL) (b) with respect to 7-bit activations and 8-bit weights and three augmentations applied to the MNIST test set: Brightness [1.5-3.5], Rotation [15°-75°] and Horizontal shift [0.1-0.5 of image size].}
\label{fig:appx_mnist_augmentations}
\end{figure*}

The additional results for MNIST-based experiments are presented in Figures~\ref{fig:appx_mnist_activation} (a-e),~\ref{fig:appx_mnist_weight} (a-e) and~\ref{fig:appx_mnist_augmentations} (a,b). Starting with the results for changing the activation precision, it can be seen that as the predictive entropy for the confusion data increases, the entropy for the test data stays constant and it increases near the smallest bit-width, as seen in Figure~\ref{fig:appx_mnist_activation} (a). As a result, the ECE, in Figure~\ref{fig:appx_mnist_activation} (b), for the test data also increases and that is due to 2 reasons: \textit{1)} The quantisation collapses when $n=3$; \textit{2)} the predictive uncertainty increases disproportionately to the error, which can be observed in SGHMC and BBB. Subsequently, the NLL on the test data stays constant and increases in the collapsed regime as seen in Figure~\ref{fig:appx_mnist_activation} (c). Nevertheless, as supported by the ECE plot for the confusion data in Figure~\ref{fig:appx_mnist_activation} (d), by becoming more uncertain, the activation quantisation reduced the confidence of BNNs on the confusion dataset, which is desired. This can be also observed on the NLL for the confusion data in Figure~\ref{fig:appx_mnist_activation} (e). Comparing these results to Figures~\ref{fig:appx_mnist_weight} (a-e), it can be seen that weight quantisation follows similar trends, and the ECE and the predictive entropy collapse, when $n$ reaches extrema. Looking in more detail at the aPE (a) and NLL (b) on test data with augmentations in Figures~\ref{fig:appx_mnist_augmentations} when $n=7$ for activations and $n=8$ for weights, it can be seen that the methods remain robust under augmentations, but at the same time they are more uncertain. This further supports the claim that the BNNs can indeed be already quantised to approximately 8-bit integer representation. In addition, we present 32-bit floating-point results with respect to Figures~\ref{fig:appx_mnist_augmentations_float} (a-d) and there are barely any visible deviations with respect to the quantised counterparts. 

\subsection{CIFAR-10}

\begin{figure*}
\centering
\begin{subfigure}{.49\textwidth}
  \centering
  \includegraphics[width=1.\linewidth]{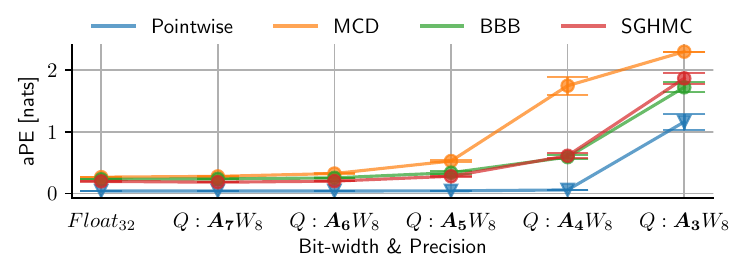}
  \caption{}
  \label{fig:appx_cifar_activation_entropy}
\end{subfigure}%
\begin{subfigure}{.49\textwidth}
  \centering
  \includegraphics[width=1.\linewidth]{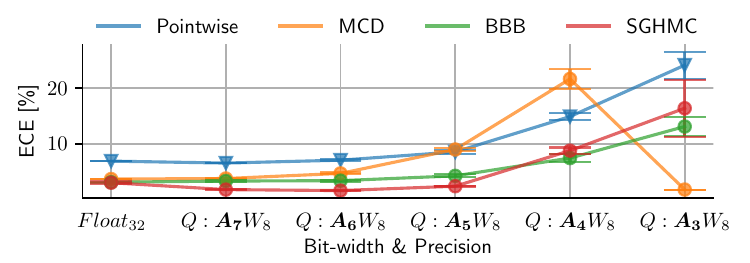}
  \caption{}
  \label{fig:appx_cifar_activation_ece}
\end{subfigure}
\newline
\begin{subfigure}{.49\textwidth}
  \centering
  \includegraphics[width=1.\linewidth]{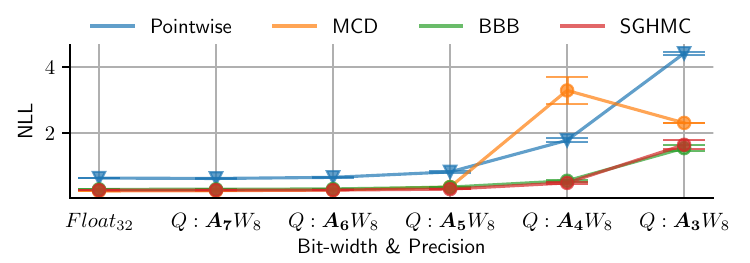}
  \caption{}
  \label{fig:appx_cifar_activation_nll}
\end{subfigure}
\begin{subfigure}{.49\textwidth}
  \centering
  \includegraphics[width=1.\linewidth]{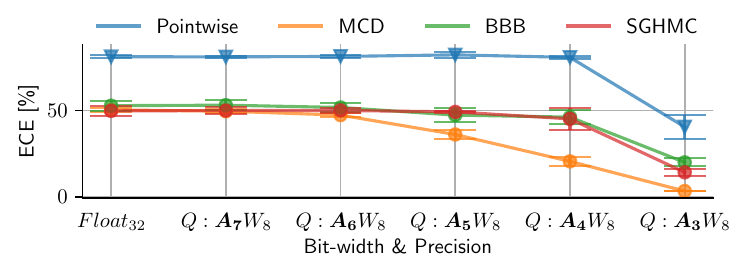}
  \caption{}
  \label{fig:appx_cifar_activation_random_ece}
\end{subfigure}
\begin{subfigure}{.49\textwidth}
  \centering
  \includegraphics[width=1.\linewidth]{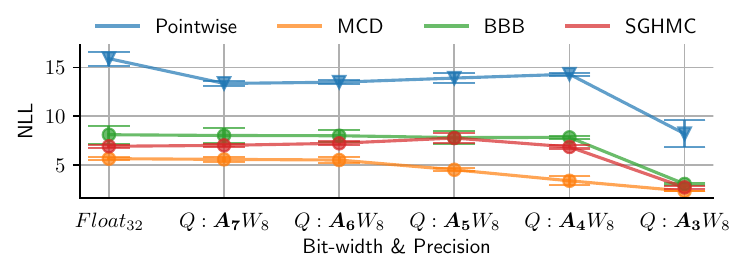}
  \caption{}
  \label{fig:appx_cifar_activation_random_nll}
\end{subfigure}
\caption{\textit{Changing activation precision, fixing weight precision.} CIFAR-10 results with respect to average predictive entropy (aPE) (a), expected calibration error (ECE) (b) and negative log-likelihood (NLL) (c) on test data and ECE (d) and NLL (e) on SVHN. Q stands for quantised activations (A) and weights (W). Subscript denotes bit-width. All methods collapse when the bit-width $\leq 4$ for A.}
\label{fig:appx_cifar_activation}
\end{figure*}

\begin{figure*}
\centering
\begin{subfigure}{.49\textwidth}
  \centering
  \includegraphics[width=1.\linewidth]{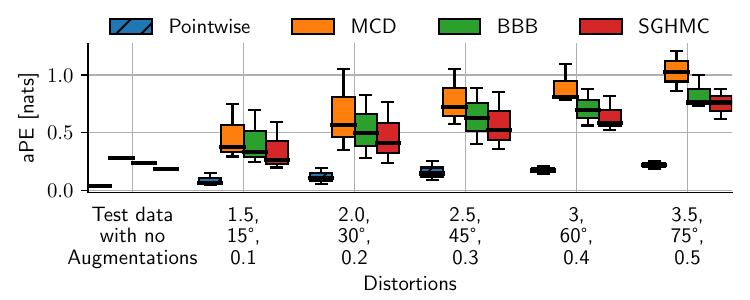}
  \caption{}
  \label{fig:appx_cifar_augmentations_ece}
\end{subfigure}
\begin{subfigure}{.49\textwidth}
  \centering
  \includegraphics[width=1.\linewidth]{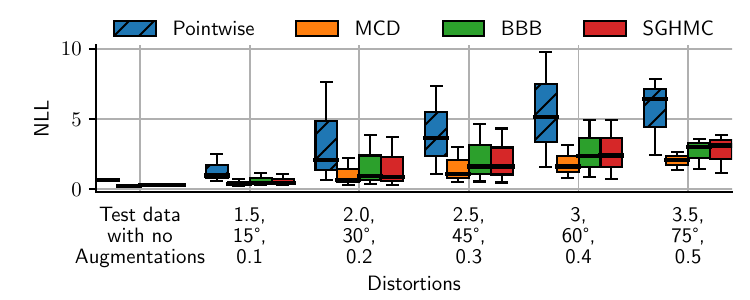}
  \caption{}
  \label{fig:appx_cifar_augmentations_nll}
\end{subfigure}
\caption{Average predictive entropy (aPE) (a) and negative log-likelihood (NLL) (b) with respect to 7-bit activations and 8-bit weights and three augmentations applied to the CIFAR-10 test set: Brightness [1.5-3.5], Rotation [15°-75°] and Horizontal shift [0.1-0.5 of image size].}
\label{fig:appx_cifar_augmentations}
\end{figure*}

The additional results for CIFAR-10-based experiments are presented in Figures~\ref{fig:appx_cifar_activation} (a-e),~\ref{fig:appx_cifar_weight} (a-e) and~\ref{fig:appx_cifar_augmentations} (a,b). Similarly to MNIST experiments, as seen in Figures~\ref{fig:appx_cifar_activation} (a-e), it can be observed that as the accuracy degrades, so does rightfully the predictive confidence, best seen in Figure~\ref{fig:appx_cifar_activation} (a). At the same time the ECE on the test data also increases, and on the confusion data it tends to decrease, as seen in Figures~\ref{fig:appx_cifar_activation} (b,d). However, this time it is due to the collapse in the activations, where the bit-width is not satisfactory and a more advanced quantisation scheme might be needed for $n \leq 4$. The results for weight quantisation support these observations as seen in Figures~\ref{fig:appx_cifar_weight} (a-e), nevertheless with smaller impact on the ranges of the deviations. This suggests a potential investigation into mixed precision representation with low bit-width of weights, while preserving the activation precision. As seen in Figure~\ref{fig:appx_cifar_augmentations} (a,b) presenting results on test data with augmentations when $n=7$ for activations and $n=8$ for weights, it can be seen that the quantisation does not prevent the Bayesian inference methods from quantifying uncertainty in their predictions. In addition, we present 32-bit floating-point results with respect to Figures~\ref{fig:appx_cifar_augmentations_float} (a-d) and there are barely any  visible deviations with respect to the quantised counterparts presented in the supplementary material or the main body of the paper.

\begin{figure*}
\centering
\begin{subfigure}{.49\textwidth}
  \centering
  \includegraphics[width=1.\linewidth]{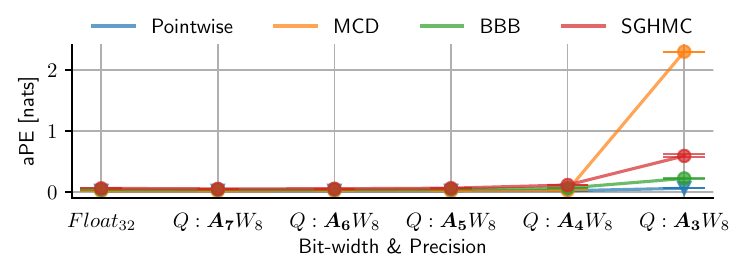}
  \caption{}
  \label{fig:appx_mnist_activation_entropy}
\end{subfigure}%
\begin{subfigure}{.49\textwidth}
  \centering
  \includegraphics[width=1.\linewidth]{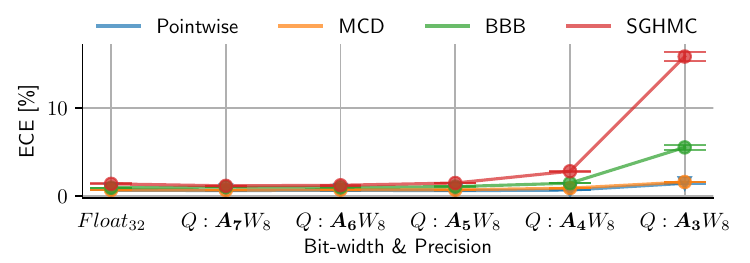}
  \caption{}
  \label{fig:appx_mnist_activation_ece}
\end{subfigure}
\end{figure*}
\begin{figure*}\ContinuedFloat
\centering
\begin{subfigure}{.49\textwidth}
  \centering
  \includegraphics[width=1.\linewidth]{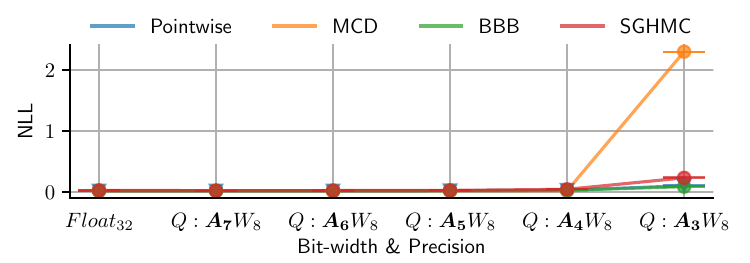}
  \caption{}
  \label{fig:appx_mnist_activation_nll}
\end{subfigure}
\begin{subfigure}{.49\textwidth}
  \centering
  \includegraphics[width=1.\linewidth]{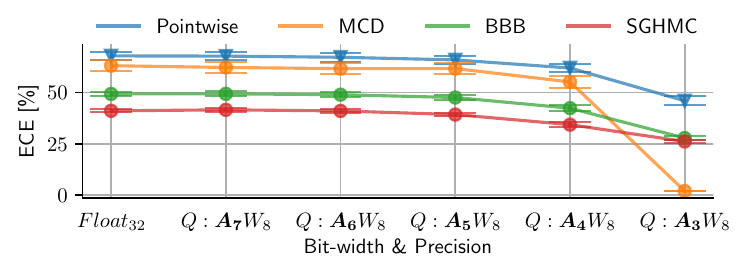}
  \caption{}
  \label{fig:appx_mnist_activation_random_ece}
\end{subfigure}
\end{figure*}
\begin{figure*}\ContinuedFloat
\centering
\begin{subfigure}{.49\textwidth}
  \centering
  \includegraphics[width=1.\linewidth]{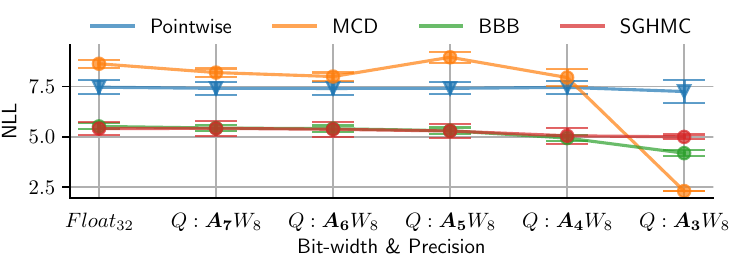}
  \caption{}
  \label{fig:appx_mnist_activation_random_nll}
\end{subfigure}
\caption{\textit{Changing activation precision, fixing weight precision.} MNIST results with respect to average predictive entropy (aPE) (a), expected calibration error (ECE) (b) and negative log-likelihood (NLL) (c) on test data and ECE (d) and NLL (e) on FashionMNIST. Q stands for quantised activations (A) and weights (W). Subscript denotes bit-width. MCD collapses when the bit-width $\leq 3$ for A.}
\label{fig:appx_mnist_activation}
\end{figure*}
\begin{figure*}
\centering
\begin{subfigure}{.49\textwidth}
  \centering
  \includegraphics[width=1.\linewidth]{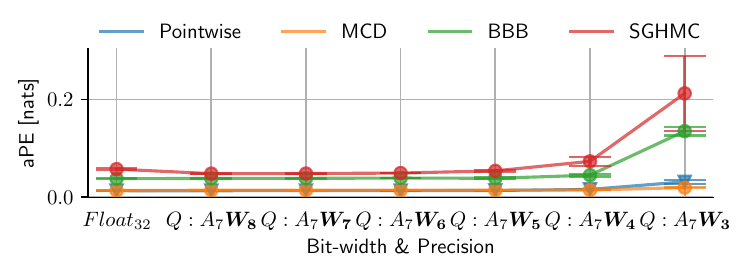}
  \caption{}
  \label{fig:appx_mnist_weight_entropy}
\end{subfigure}%
\begin{subfigure}{.49\textwidth}
  \centering
  \includegraphics[width=1.\linewidth]{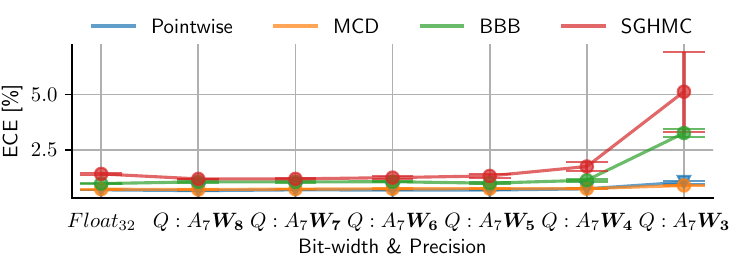}
  \caption{}
  \label{fig:appx_mnist_weight_ece}
\end{subfigure}
\begin{subfigure}{.49\textwidth}
  \centering
  \includegraphics[width=1.\linewidth]{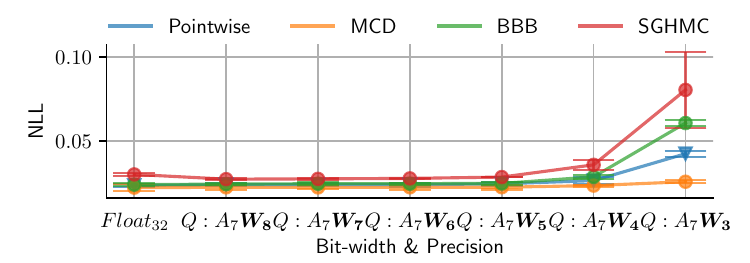}
  \caption{}
  \label{fig:appx_mnist_weight_nll}
\end{subfigure}
\begin{subfigure}{.49\textwidth}
  \centering
  \includegraphics[width=1.\linewidth]{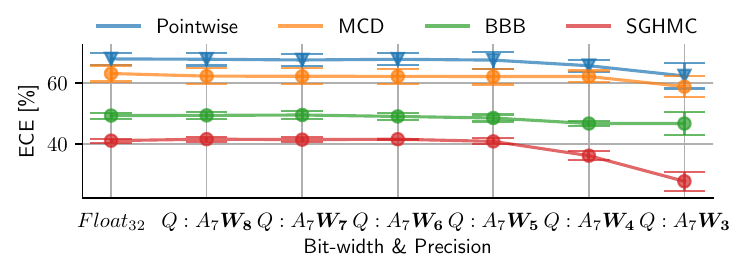}
  \caption{}
  \label{fig:appx_mnist_weight_random_ece}
\end{subfigure}
\end{figure*}
\begin{figure*}\ContinuedFloat
\centering
\begin{subfigure}{.49\textwidth}
  \centering
  \includegraphics[width=1.\linewidth]{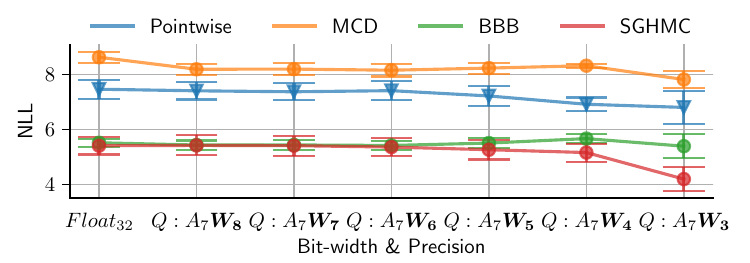}
  \caption{}
  \label{fig:appx_nll_weight_random_nll}
\end{subfigure}
\caption{\textit{Fixing activation precision, changing weight precision.} MNIST results with respect to average predictive entropy (aPE) (a), expected calibration error (ECE) (b) and negative log-likelihood (NLL) (c) on test data and ECE (d) and NLL (e) on FashionMNIST data. Q stands for quantised activations (A) and weights (W). Subscript denotes bit-width.}
\label{fig:appx_mnist_weight}
\end{figure*}
\begin{figure*}
\centering
\begin{subfigure}{.5\textwidth}
  \centering
  \includegraphics[width=1.\linewidth]{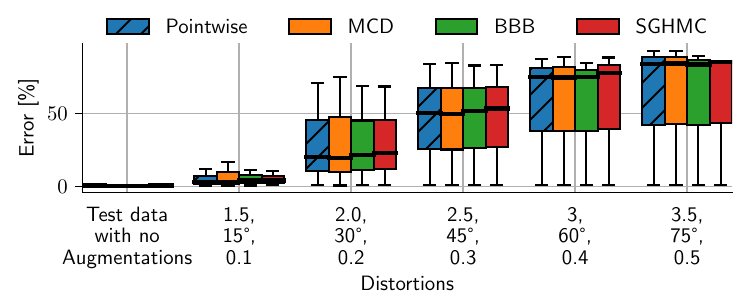}
  \caption{}
  \label{fig:appx_mnist_augmentations_error_float}
\end{subfigure}%
\begin{subfigure}{.5\textwidth}
  \centering
  \includegraphics[width=1.\linewidth]{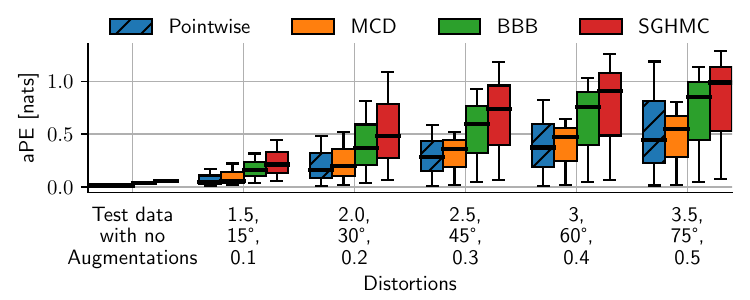}
  \caption{}
  \label{fig:appx_mnist_augmentations_entropy_float}
\end{subfigure}
\end{figure*}
\begin{figure*}\ContinuedFloat
\centering
\begin{subfigure}{.49\textwidth}
  \centering
  \includegraphics[width=1.\linewidth]{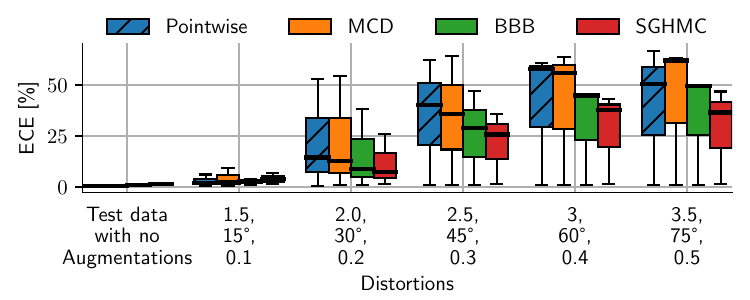}
  \caption{}
  \label{fig:appx_mnist_augmentations_ece_float}
\end{subfigure}
\begin{subfigure}{.49\textwidth}
  \centering
  \includegraphics[width=1.\linewidth]{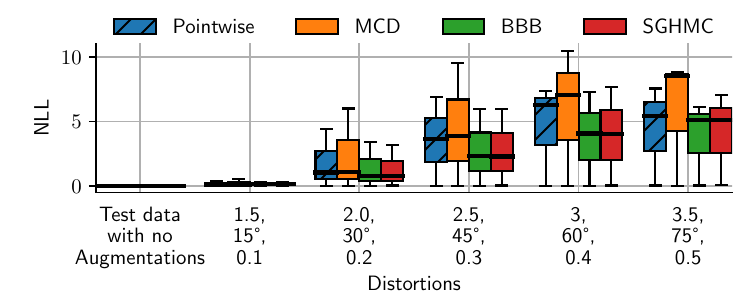}
  \caption{}
  \label{fig:appx_mnist_augmentations_nll_float}
\end{subfigure}
\caption{Classification error (a), average predictive entropy (aPE) (b), expected calibration error (ECE) (c) and negative log-likelihood (NLL) (d) with respect to 32-bit floating-point activations and weights and three augmentations applied to the MNIST test set: Brightness [1.5-3.5], Rotation [15°-75°] and Horizontal shift [0.1-0.5 of image size].}
\label{fig:appx_mnist_augmentations_float}
\end{figure*}
\begin{figure*}
\centering
\begin{subfigure}{.49\textwidth}
  \centering
  \includegraphics[width=1.\linewidth]{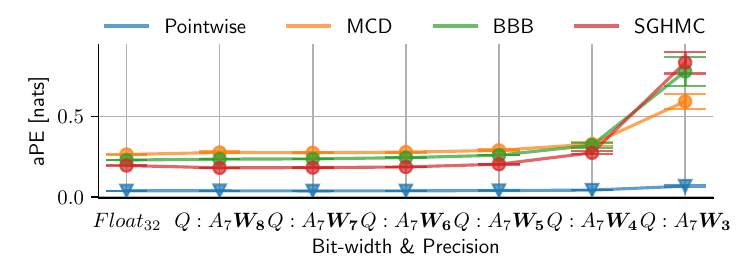}
  \caption{}
  \label{fig:appx_cifar_weight_entropy}
\end{subfigure}%
\begin{subfigure}{.49\textwidth}
  \centering
  \includegraphics[width=1.\linewidth]{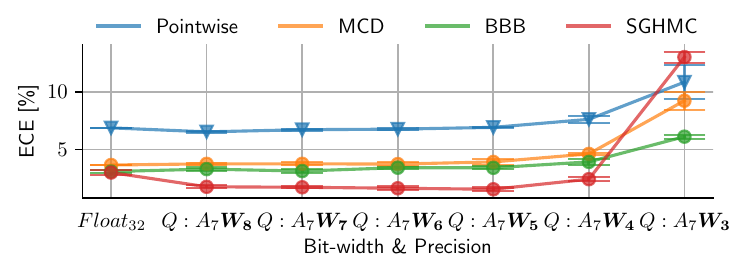}
  \caption{}
  \label{fig:appx_cifar_weight_ece}
\end{subfigure}
\end{figure*}
\begin{figure*}\ContinuedFloat
\centering
\begin{subfigure}{.49\textwidth}
  \centering
  \includegraphics[width=1.\linewidth]{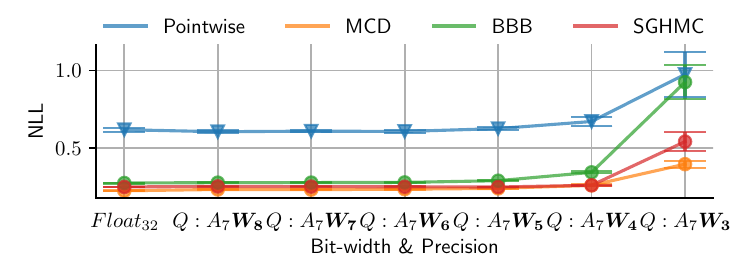}
  \caption{}
  \label{fig:appx_cifar_weight_nll}
\end{subfigure}
\begin{subfigure}{.49\textwidth}
  \centering
  \includegraphics[width=1.\linewidth]{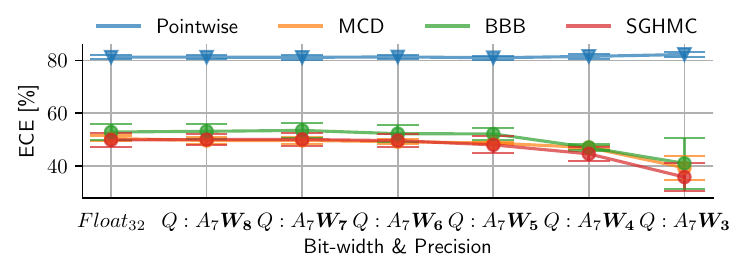}
  \caption{}
  \label{fig:appx_cifar_weight_random_ece}
\end{subfigure}
\end{figure*}
\begin{figure*}\ContinuedFloat
\centering
\begin{subfigure}{.49\textwidth}
  \centering
  \includegraphics[width=1.\linewidth]{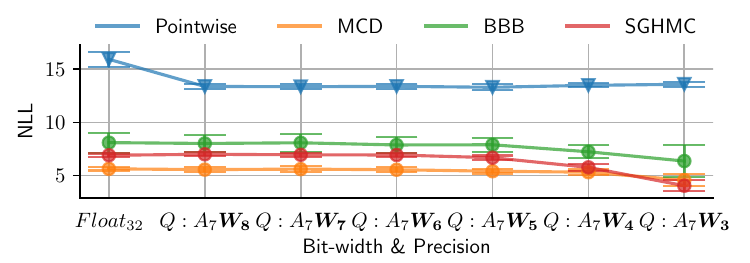}
  \caption{}
  \label{fig:appx_cifar_weight_random_nll}
\end{subfigure}

\caption{\textit{Fixing activation precision, changing weight precision.} CIFAR-10 results with respect to average predictive entropy (aPE) (a), expected calibration error (ECE) (b) and negative log-likelihood (NLL) (c) on test data and ECE (d) and NLL (e) on SVHN. Q stands for quantised activations (A) and weights (W). Subscript denotes bit-width. All methods except MCD collapse when the bit-width $\leq 3$ for W.}
\label{fig:appx_cifar_weight}
\end{figure*}
\begin{figure*}
\centering
\begin{subfigure}{.49\textwidth}
  \centering
  \includegraphics[width=1.\linewidth]{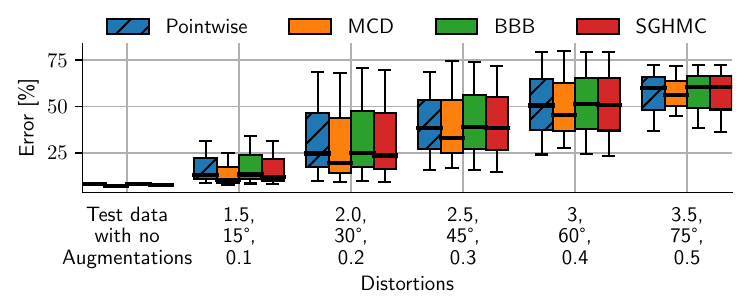}
  \caption{}
  \label{fig:appx_cifar_augmentations_error_float}
\end{subfigure}%
\begin{subfigure}{.49\textwidth}
  \centering
  \includegraphics[width=1.\linewidth]{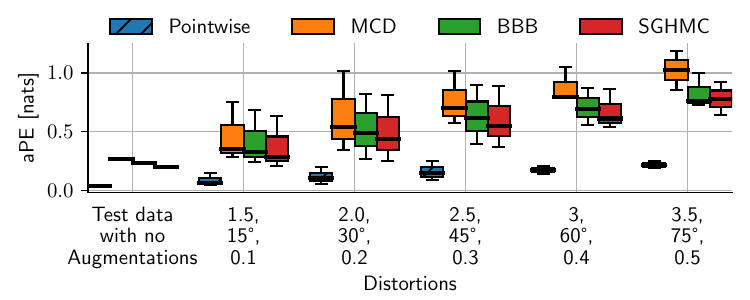}
  \caption{}
  \label{fig:appx_cifar_augmentations_entropy_float}
\end{subfigure}
\end{figure*}
\begin{figure*}\ContinuedFloat
\centering
\begin{subfigure}{.49\textwidth}
  \centering
  \includegraphics[width=1.\linewidth]{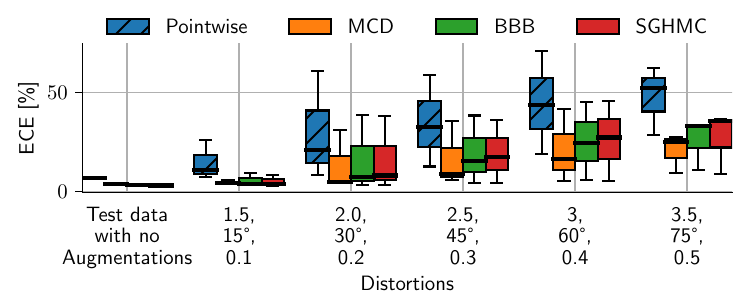}
  \caption{}
  \label{fig:appx_cifar_augmentations_ece_float}
\end{subfigure}
\begin{subfigure}{.49\textwidth}
  \centering
  \includegraphics[width=1.\linewidth]{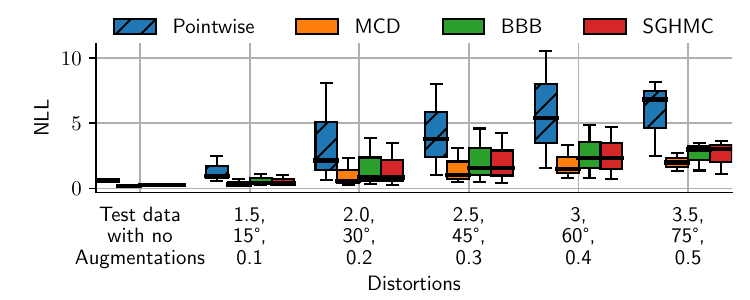}
  \caption{}
  \label{fig:appx_cifar_augmentations_nll_float}
\end{subfigure}
\caption{Classification error (a), average predictive entropy (aPE) (b) and expected calibration error (ECE) (c) and negative log-likelihood (NLL) (d) with respect to 32-bit floating-point activations and weights and three augmentations applied to the CIFAR-10 test set: Brightness [1.5-3.5], Rotation [15°-75°] and Horizontal shift [0.1-0.5 of image size].}
\label{fig:appx_cifar_augmentations_float}
\end{figure*}

\end{document}